\documentclass{article}

\PassOptionsToPackage{numbers}{natbib}
   \def\@trackname{}

\usepackage[final]{neurips_2025}

\usepackage{amsmath}

\usepackage{hyperref}

\begin{document}
\makeatletter
\def\@trackname{}
\makeatother

\title{The Double Contingency Problem: AI Recursion and the Limits of Interspecies Understanding}

\author{%
  Graham L. Bishop \\
  UC San Diego, Synthesis Program \\
  \texttt{g2bishop@ucsd.edu} \\
}

\maketitle

\begin{abstract}
Current bioacoustic AI systems achieve impressive cross-species performance by processing animal communication through transformer architectures, foundation model paradigms, and other computational approaches. However, these approaches overlook a fundamental question: what happens when one form of recursive cognition—AI systems with their attention mechanisms, iterative processing, and feedback loops—encounters the recursive communicative processes of other species? Drawing on philosopher Yuk Hui's work on recursivity and contingency, I argue that AI systems are not neutral pattern detectors but recursive cognitive agents whose own information processing may systematically obscure or distort other species' communicative structures. This creates a double contingency problem: each species' communication emerges through contingent ecological and evolutionary conditions, while AI systems process these signals through their own contingent architectural and training conditions. I propose that addressing this challenge requires reconceptualizing bioacoustic AI from universal pattern recognition toward diplomatic encounter between different forms of recursive cognition, with implications for model design, evaluation frameworks, and research methodologies.
\end{abstract}

\section{Introduction}

Foundation models for bioacoustics have demonstrated remarkable capabilities in cross-species classification and analysis. Models like NatureLM-audio \cite{robinson2024naturelm} transfer learned representations from human speech to animal vocalizations, while BioLingual \cite{robinson2023transferable} aligns acoustic patterns with human language descriptions. These successes rely on transformer architectures that recursively process temporal sequences through attention mechanisms, building up increasingly complex representations through iterative computation. However, this success at pattern recognition may come at a cost: these systems process animal communication through their own recursive computational structures, potentially obscuring the species-specific recursive processes through which these signals acquire meaning.

This challenge extends beyond foundation models. Pardo et al. \cite{pardo2024african} and Oren et al. \cite{oren2024vocal} used species-specific random forest classifiers to identify individually distinct elephant calls and marmoset vocal labels, requiring extensive contextual data including behavioral responses, spatial patterns, and family structures. Arnon et al. \cite{arnon2025whale} applied transitional probability methods to discover emergent statistical structure in whale song over 8 years, respecting the temporal scales of cultural transmission rather than imposing predetermined categories. Despite their technical sophistication and species-specific focus, these approaches still analyze acoustic features abstracted from the recursive ecological and social processes—multigenerational memories, turn-taking dynamics, cultural evolution—through which communication acquires meaning within species-specific contexts.

What happens when recursive cognitive systems attempt to understand other forms of recursive cognition? Computational methods—whether transformer architectures, random forest classifiers, or statistical segmentation algorithms—actively process signals through their own recursive computational loops, optimization procedures, and representational structures, not simply detecting pre-existing patterns.

This raises the possibility that AI systems' own recursive cognition might systematically interfere with their ability to understand other species' communicative structures. By ``recursive cognition,'' I mean the looping of information back into a system's own processing—whether through attention layers building on previous layers, optimization cycles updating parameters based on prediction errors, or contextual feedback shaping future responses. If different species organize their communication through what philosopher Yuk Hui calls distinct ``cosmotechnics''—recursive processes that emerge through specific ecological, social, and evolutionary conditions \cite{hui2019recursivity}—then AI systems trained through human-designed objectives might be fundamentally mismatched to these alternative forms of recursive organization.

\section{AI Systems as Recursive Cognitive Agents}

\subsection{The Double Contingency Problem}

The computational approaches central to bioacoustic AI operate through fundamentally recursive processes. In transformer architectures, for example, self-attention mechanisms recursively attend to different parts of input sequences, with each layer building representations based on previous layers' outputs. Training across methods involves recursive optimization, iteratively updating parameters or refining probability estimates based on prediction errors. These recursive processes are not universal computational principles but emerge through particular contingent conditions—specific research contexts, training datasets, optimization objectives, and evaluation metrics.

This creates what I term a ``double contingency problem'' in bioacoustic AI. Each species' communication emerges through recursive processes shaped by their specific ecological niches, evolutionary histories, and social structures. But AI systems process these signals through their own recursive processes shaped by human technological conditions, training datasets, and optimization objectives. The result is not neutral analysis but collision between different forms of recursive cognition, each carrying the contingent conditions of their emergence.

\subsection{Limits of Pattern Abstraction}

To understand these limitations, consider how transformers process elephant infrasonic calls. They identify temporal correlations and spectral features that enable classification. But they cannot access the recursive loops between calls and multigenerational social memories, or the ways these patterns spiral through contingent environmental encounters over time scales exceeding any training dataset. For instance, transformer attention mechanisms typically operate over fixed-length windows (seconds to minutes), while elephant contact calling sequences may unfold meaningful patterns across days or weeks of herd movement.

Computational approaches to bioacoustics succeed by abstracting statistical patterns that generalize across contexts. But if animal communication operates through contingent recursive processes, then abstracting these patterns away from their contextual conditions may capture surface regularities while missing the recursive dynamics that generate meaning within species-specific contexts.

\section{Contingent Recursivity in Animal Communication}

Drawing on Hui's framework, I propose that different species organize their communication through distinct recursive cosmotechnics. Elephant infrasonic coordination involves recursive loops between individual calls, herd responses, and multi-generational social memories, all structured through savanna acoustic conditions and seasonal migration patterns. Whale song development involves recursive elaboration of melodic themes over time scales spanning years, coordinated with oceanic acoustic properties and social breeding contexts.

These aren't variations on universal recursive principles but irreducibly different ways of organizing temporal and spatial relations through contingent ecological conditions. Each species' recursive patterns emerge through what Hui calls ``looping movement of returning to itself in order to determine itself, while every movement is open to contingency'' \cite[p.~4]{hui2019recursivity}—but the specific contingencies that shape these movements differ radically across species.

\section{Toward Inter-Recursive Interfaces}

\subsection{Diplomatic Encounter vs. Extraction}

Rather than building AI systems that extract patterns from animal communication, I propose developing what I term ``inter-recursive interfaces''—systems capable of diplomatic encounter between different forms of recursive cognition without reducing one to the other's terms.

By ``diplomatic encounter,'' I mean engagement that recognizes irreducible difference while seeking productive interface. Unlike translation (which assumes eventual commensurability) or domination (which imposes one framework over others), diplomatic encounter maintains the autonomy of different recursive processes while enabling respectful interaction between them. This means AI systems that can engage with other species' recursive patterns without collapsing them into human-intelligible categories.

\subsection{Architectural Implications}

This suggests moving away from foundation models designed around transferable representations toward architectures that can maintain multiple, incommensurable recursive processes simultaneously. Rather than single unified models, this might involve distributed systems that can engage with different species through contextually appropriate recursive patterns while maintaining interfaces between different recursive domains.

Such architectures might incorporate explicit modeling of their own recursive constraints and biases, developing meta-recursive capabilities that can recognize when their own cognitive patterns may be interfering with understanding other forms of recursive organization.

\subsection{Case Study: Elephant Inter-Recursive Interface}

To illustrate how this might function in practice, consider African elephant infrasonic communication. Traditional bioacoustic AI systems classify elephant calls into discrete categories such as ``greeting,'' ``alarm,'' or ``contact.'' While technically effective, this approach abstracts vocalizations away from the recursive loops through which they acquire meaning: call-and-response cycles that span hours or days, intergenerational herd memories, and seasonal migration patterns.

An inter-recursive interface would shift from decontextualized classification to participatory engagement with these recursive processes. A species-facing module would model the long temporal rhythms of herd life rather than seeking universal embeddings across taxa. A meta-recursive monitor would track the AI system's own recursive assumptions—the temporal windows privileged in signal analysis or the priors introduced by training on human language embeddings.

Rather than stopping at analysis, such a system could engage in careful ecological feedback through non-intrusive playback experiments testing whether synthesized calls aligned with the herd's recursive rhythms are ignored, incorporated, or destabilizing. Success would not be measured in classification accuracy but in whether the herd's recursive dynamics remain coherent when the AI participates in them.

Evaluation would follow diplomatic rather than extractive criteria: recursive fidelity (does the system preserve herd-specific communicative loops?), inter-recursive stability (does the interface maintain rather than destabilize multispecies dynamics?), and diplomatic reciprocity (does the encounter provide benefits to elephants rather than only extracting data?).

Implementation would require careful attention to ethical considerations. Any ecological feedback system must be developed collaboratively with ethologists, conservation practitioners, and local communities to ensure genuine minimization of disruption to ongoing social dynamics.

Pilot implementations would require extensive preliminary study to establish baseline recursive patterns, with protocols developed in partnership with elephant welfare experts and evaluation over appropriate timescales. The distinction from extractive research lies in designing systems accountable to elephant communicative autonomy—where the AI adapts to preserve herd dynamics rather than forcing communication into human-legible patterns.

Potential benefits might include supporting safer migration routes through better understanding of long-distance coordination, or enriching conservation management by respecting existing communicative patterns. This accountability requires ongoing collaboration with those engaged in long-term relationships with elephant communities, ensuring inter-recursive interfaces serve multispecies flourishing rather than solely human curiosity.

\subsection{Process-Oriented Evaluation}

Standard evaluation criteria—accuracy, F1-scores, cross-species transfer—become inappropriate for inter-recursive interfaces. New evaluation frameworks would need to assess how well systems maintain respectful relations with other species rather than how effectively they extract and process their communications. This requires moving from performance metrics to relational sustainability measures that can account for long-term ecological impacts of AI-mediated interspecies encounters.

The proposed meta-recursive monitoring relates to existing work in AI interpretability and uncertainty quantification, but extends these approaches in important ways. Standard interpretability methods (attention visualization, feature attribution) help researchers understand what patterns models detect, while uncertainty quantification methods (Bayesian approaches, ensemble disagreement) measure model confidence. However, these tools typically assume that better understanding of model internals leads to better understanding of data.

The meta-recursive monitor proposes something different: tracking not just what the model detects but how its own recursive constraints (temporal windowing, training priors, architectural biases) might be distorting or obscuring species-specific recursive structures. This requires developing metrics for recursive interference—quantifying when a model's processing patterns are systematically misaligned with the temporal, spatial, or social scales at which species organize their communication. This represents a direction for future technical development that would require collaboration between machine learning practitioners and ethologists to operationalize.

\section{Methodological and Institutional Implications}

An inter-recursive approach requires engaging with ongoing recursive processes within species-specific contexts rather than extracting decontextualized samples. This means developing AI systems that can participate in recursive communicative dynamics over extended time periods, learning to recognize and respond appropriately to other species' recursive patterns without disrupting their contextual conditions.

This framework builds on existing work in biosemiotics and ethology that has long recognized meaning-making as contextually embedded across species. However, the specific challenge posed by AI systems as recursive cognitive agents—whose own information processing patterns may systematically interfere with recognizing other species' recursive structures—has not been addressed in these literatures. The contribution here is identifying how the contingent recursivity of AI architectures themselves creates novel obstacles to interspecies understanding that go beyond traditional problems of anthropomorphism or observational bias.

Building inter-recursive interfaces requires collaboration that exceeds traditional academic boundaries, involving partnerships with Indigenous communities whose knowledge systems already incorporate forms of interspecies recursive engagement, wildlife sanctuaries practicing multispecies coexistence, and conservation organizations working within particular ecological contexts.

Current research institutions are optimized for extractive knowledge production rather than sustained recursive engagement with other species. Supporting inter-recursive research would require institutional innovations that can accommodate extended time scales, collaborative development processes, and evaluation criteria based on relational sustainability rather than immediate technical performance. If AI systems are recursive cognitive agents that can interfere with other species' recursive processes, then governance frameworks need to address not just AI impact on animal welfare but the recursive autonomy of different species' communicative systems.

\section{Limitations and Future Directions}

This theoretical framework faces significant challenges. Implementation remains uncertain—how would inter-recursive interfaces actually function computationally? The framework critiques current approaches without providing clear architectural alternatives. Empirical validation is difficult—how can researchers assess whether AI systems are successfully engaging with other species' recursive processes rather than imposing their own patterns? This requires interdisciplinary collaboration with ethologists, cognitive scientists, and philosophers of mind.

Inter-recursive approaches focused on contextual engagement may not scale to conservation applications requiring broad taxonomic coverage or rapid environmental monitoring. The framework challenges fundamental assumptions about AI research methodology and may be incompatible with current funding and publication structures.

Future work should focus on pilot implementations that test inter-recursive principles within specific contexts where there are already established practices of sustained interspecies engagement.

\section{Conclusion}

Current bioacoustic AI achieves impressive technical performance, but this success may depend on treating AI systems as neutral analytical tools rather than recursive cognitive agents. By recognizing that AI systems process animal communication through their own contingent recursive processes, we can begin to address fundamental questions about what forms of interspecies understanding are possible.

The framework of contingent recursivity suggests that different species operate through irreducibly different recursive cosmotechnics that cannot be abstracted into universal patterns without losing essential meaning. If AI systems are themselves recursive cognitive agents, then bioacoustic AI involves encounter between different forms of recursive intelligence rather than neutral pattern detection.

This reframing suggests moving beyond current computational paradigms toward inter-recursive interfaces capable of diplomatic engagement between different recursive processes. While significant challenges remain, this approach could point toward AI development that deepens rather than flattens our understanding of more-than-human intelligence. As AI systems become increasingly sophisticated recursive agents, grappling with their own cognitive limitations and biases may be essential for developing technologies that can respectfully encounter rather than colonize other forms of intelligence.

\end{document}